\documentclass[conference]{IEEEtran}
\IEEEoverridecommandlockouts
\usepackage{cite}
\usepackage{amsmath,amssymb,amsfonts}
\usepackage{algorithmic}
\usepackage{textcomp}
\usepackage[table,xcdraw,dvipsnames]{xcolor}
\definecolor{lightpink1}{HTML}{FFD5D9}
\definecolor{lightpink2}{HTML}{FFE2E6}
\definecolor{lightblue1}{HTML}{D6ECFF}  
\definecolor{lightblue2}{HTML}{E9F4FF}  
\usepackage{url}
\usepackage{stfloats}
\usepackage{float} 
\usepackage[hidelinks, colorlinks=true, linkcolor=blue, citecolor=blue, breaklinks=true]{hyperref}
\usepackage[T1]{fontenc}
\usepackage{graphicx}
\usepackage{multirow}
\usepackage{makecell}
\usepackage{colortbl}
\usepackage{booktabs}
\usepackage{pifont}
\usepackage{dsfont}
\usepackage{subfigure}
\usepackage{subcaption}
\usepackage{stfloats}
\usepackage{marvosym}
\usepackage[utf8]{inputenc}
\def\BibTeX{{\rm B\kern-.05em{\sc i\kern-.025em b}\kern-.08em
    T\kern-.1667em\lower.7ex\hbox{E}\kern-.125emX}}

\title{HPR-SAM: Hierarchical Probabilistic Representation Learning for Prompt-free SAM-based Medical Image Segmentation
}

\author{

\IEEEauthorblockN{
Yingzhen Hu\textsuperscript{2},
Yiheng Zhong\textsuperscript{1},
Keying Zhu\textsuperscript{2},
Zimu Zhang\textsuperscript{2},\\
Zihan Ye\textsuperscript{2},
Sifan Song\textsuperscript{2,\dag},
Jionglong Su\textsuperscript{2,\dag},
Xiaofeng Liu\textsuperscript{1,\dag}
}

\IEEEauthorblockA{
\textsuperscript{1}Yale University, United States\\
\textsuperscript{2}Xi'an Jiaotong-Liverpool University, China\\
Emails: sifan.song@xjtlu.edu.cn, jionglong.su@xjtlu.edu.cn, xiaofeng.liu@yale.edu
}

\thanks{\textsuperscript{\dag}Corresponding authors.}
}

\begin{document}
\maketitle

\begin{abstract}
Prompt-free adaptation of the Segment Anything Model (SAM) has emerged as a promising paradigm for automatic medical image segmentation. Existing methods mainly focus on prompt generation, while overlooking that prompt quality is fundamentally constrained by the expressiveness of anatomical representations. However, deterministic prototypes or semantic tokens are insufficient to jointly capture global anatomical priors, intra-structure diversity, and local structural reliability. To address this limitation, we propose the Hierarchical Probabilistic Representation (HPR) framework, which learns complementary anatomical representations through Distributional Anatomical Representation (DAR), Multi-component Anatomical Representation (MAR), and Local Reliability Representation (LRR), and integrates their predictions via Hierarchical Prediction Fusion (HPF) while remaining compatible with the original SAM decoder. Experiments on the Synapse, LA, and PROMISE12 datasets demonstrate that HPR-SAM achieves state-of-the-art performance on Synapse and the best performance under few-shot settings on LA and PROMISE12, validating the effectiveness of the proposed hierarchical probabilistic representation learning framework for prompt-free medical image segmentation. Our anonymous code is released at
\url{https://anonymous.4open.science/r/HPR-SAM-E4AF}.
\end{abstract}


\section{Introduction}
\label{sec:intro}

Medical image segmentation is a fundamental task in medical image analysis and plays an essential role in computer-aided diagnosis, treatment planning, and surgical navigation~\cite{ronneberger2015unet,isensee2021nnu}. Recently, the Segment Anything Model (SAM)~\cite{kirillov2023segment} introduced a prompt-driven segmentation paradigm with remarkable generalization capability, inspiring extensive studies on adapting foundation models to medical image segmentation~\cite{ma2024segment,zhang2023customized,wu2025adapting}. Subsequent studies have demonstrated that SAM can be effectively adapted to medical images through domain adaptation and parameter-efficient fine-tuning, providing a promising foundation for universal medical image segmentation~\cite{ma2024segment,zhang2023customized,cheng2024unleashing}.

Despite these advances, adapting SAM to fully automatic medical image segmentation remains challenging. The original SAM relies on manually provided point or box prompts, which increases human interaction and limits its clinical practicality. To eliminate manual intervention, recent studies have explored prompt-free SAM by automatically generating dense or sparse prompts while preserving SAM's original prompt-to-mask decoding paradigm~\cite{hu2023efficiently,zhang2023customized,cheng2024unleashing}. Existing approaches mainly focus on improving automatic prompt learning, enabling SAM to perform segmentation without manual prompts~\cite{yan2025pgpsam,yue2024surgicalsam,wei2024dapsam}.
\begin{figure}[t]
    \centering
    \includegraphics[width=\linewidth]{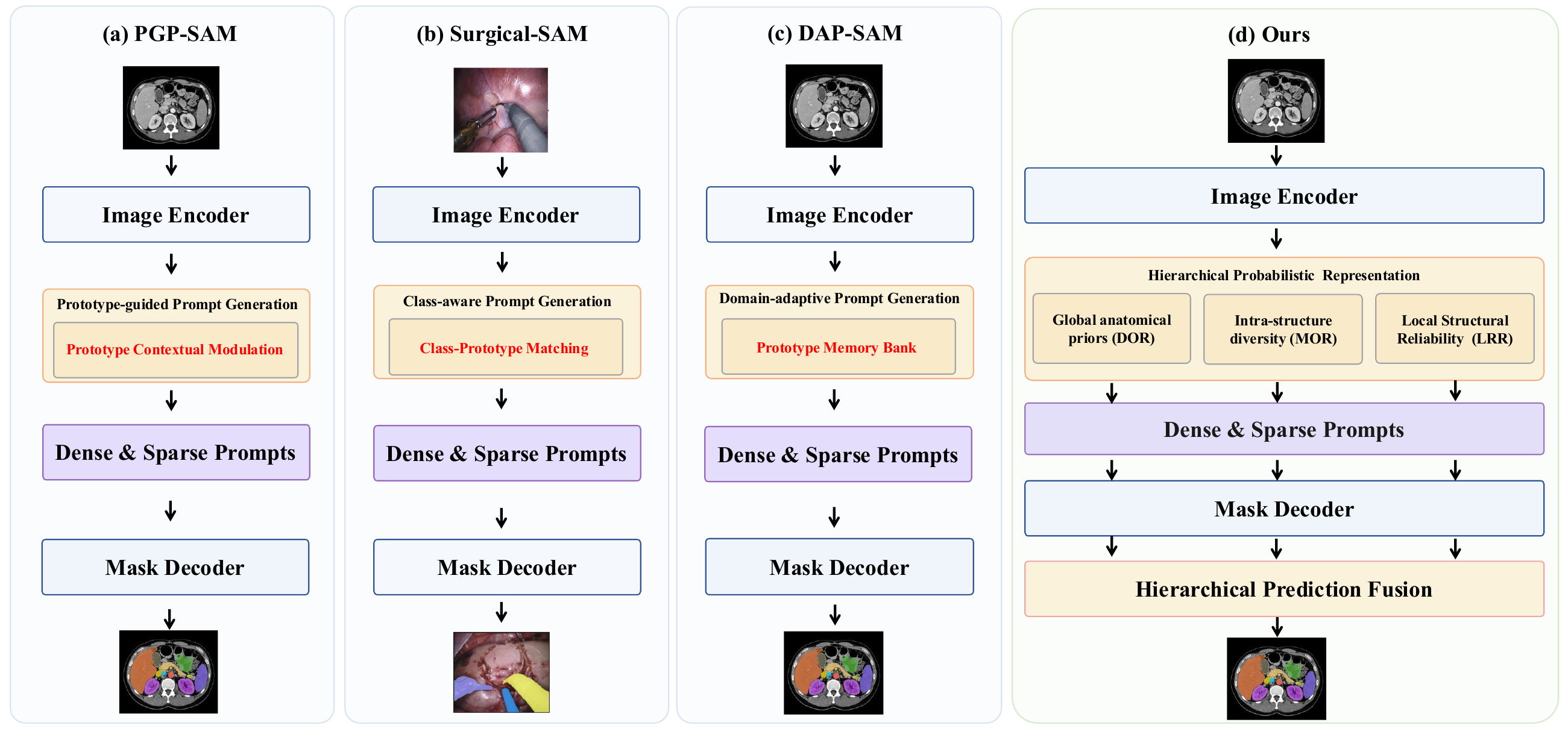}
    \caption{Comparison of representative prompt-free SAM frameworks.
(a) PGP-SAM employs prototype contextual modulation.
(b) SurgicalSAM performs class-prototype matching.
(c) DAP-SAM adopts prototype memory banks for domain adaptation.
(d) HPR-SAM learns hierarchical probabilistic representations through DAR, MAR, and LRR, followed by HPF for adaptive prediction fusion.}
    \label{fig1}
\end{figure}

Recent studies have substantially advanced automatic prompt learning. As illustrated in Fig.~\ref{fig1}, representative prompt-free SAM methods, including PGP-SAM~\cite{yan2025pgpsam}, SurgicalSAM~\cite{yue2024surgicalsam}, and DAP-SAM~\cite{wei2024dapsam}, improve automatic prompt learning through different prompt generation strategies, such as prototype-guided prompt generation, Class-aware prompt generation, and domain-adaptive prompt generation, respectively. Although these methods adopt different implementations, they share a common design philosophy: improving automatic prompts by designing increasingly sophisticated prompt generation modules while largely relying on deterministic anatomical representations learned from image features.

 In prompt-free SAM adaptation, automatic prompts are not independent learnable entities. Instead, they are obtained by projecting anatomical representations learned from image observations into the prompt space. Consequently, the anatomical information carried by prompts is fundamentally constrained by the expressive capability of these representations. If the underlying anatomical representation fails to faithfully characterize anatomical structures, even a sophisticated prompt generator can hardly recover the missing anatomical semantics. Therefore, the key lies in learning expressive anatomical representations before prompt generation.

This naturally raises the question: What constitutes an expressive anatomical representation for automatic prompt generation? We argue that a complete representation should jointly capture global anatomical priors, intra-structure diversity, and local structural reliability. However, as shown in Fig.~\ref{fig1}, existing prompt-free SAM methods often compress anatomical targets into deterministic prototype-based representations, including (a) prototype contextual modulation, (b) class-prototype matching, and (c) prototype memory banks. Such representations are insufficient to jointly model the above complementary properties. Consequently, improving prompt generation alone is insufficient, as the learned anatomical representations remain incomplete~\cite{yan2025pgpsam,yue2024surgicalsam,wei2024dapsam}.

\begin{figure}[t]
    \centering
    \includegraphics[width=\columnwidth]{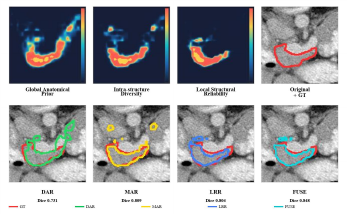}
   \caption{
Visualization of the three complementary anatomical properties. The first three columns visualize their corresponding response heatmaps, and the remaining columns present the corresponding segmentation results together with the final fused prediction.
}
    \label{fig:ablation_visualization}
\end{figure}
Inspired by recent advances in probabilistic representation learning, we revisit prompt-free SAM from the perspective of anatomical representation learning rather than prompt generator design and propose a Hierarchical Probabilistic Representation (HPR) framework. Instead of directly optimizing prompt generation, HPR learns hierarchical probabilistic representations from image tokens and subsequently projects them into SAM-compatible dense and sparse prompts. Specifically, HPR models global anatomical priors, intra-structure anatomical diversity, and local structural reliability through three probabilistic representation learning modules, namely Distributional Anatomical Representation (DAR), Multi-component Anatomical Representation (MAR), and Local Reliability Representation (LRR), respectively. As illustrated in Fig.~\ref{fig:ablation_visualization}, these three representations provide complementary cues for anatomical representation. Global anatomical priors provide stable anatomical-level localization, intra-structure diversity captures morphological variations to improve structural completeness and produce clearer boundaries, while local structural reliability emphasizes trustworthy local evidence, suppressing unreliable responses in ambiguous and low-contrast regions. Faithfully modeling these complementary anatomical properties leads to a richer anatomical representation for automatic prompt generation. Finally, the learned representations are projected into SAM-compatible dense and sparse prompts and decoded into three complementary prediction maps, which are adaptively integrated by Hierarchical Prediction Fusion (HPF) to produce the final segmentation.

The contributions of this work are summarized below:

\begin{itemize}

\item To the best of our knowledge, this is the first work to revisit automatic prompt learning from the perspective of anatomical representation learning, revealing that prompt quality is constrained by representation expressiveness rather than prompt generator design alone.

\item We propose Hierarchical Probabilistic  Representation (HPR) framework, which learns anatomically complete representations by jointly modeling global anatomical priors, intra-structure anatomical diversity, and local structural reliability.

\item The proposed DAR, MAR, LRR, and HPF modules enable complementary probabilistic representation learning and adaptive prediction fusion while remaining compatible with the original SAM decoder.

\item Experiments on the Synapse, LA, and PROMISE12 datasets demonstrate that HPR-SAM achieves state-of-the-art performance across diverse medical image segmentation benchmarks.
\end{itemize}

\section{Related Work}

\subsection{SAM-based Medical Image Segmentation}

Medical image segmentation has achieved remarkable progress with convolutional neural networks (CNNs) and Transformer-based architectures. Representative methods, including U-Net~\cite{ronneberger2015unet}, nnU-Net~\cite{isensee2021nnu}, TransUNet~\cite{chen2021transunet}, UNETR~\cite{hatamizadeh2022unetr}, and Swin-Unet~\cite{cao2022swin}, have achieved excellent performance on various medical image segmentation benchmarks. However, they are typically designed for specific anatomical targets, imaging modalities, or datasets, limiting their generalization across diverse clinical scenarios.

Recently, the Segment Anything Model (SAM)~\cite{kirillov2023segment} introduced a prompt-driven segmentation paradigm, providing a new foundation for medical image segmentation. MedSAM~\cite{ma2024segment} adapts SAM to the medical domain through large-scale medical image fine-tuning, while SAM-Med2D~\cite{cheng2023sam} extends SAM to large-scale 2D medical images. Parameter-efficient methods such as SAMed~\cite{zhang2023customized} employ lightweight trainable parameters for medical adaptation, and H-SAM~\cite{cheng2024unleashing} further improves prompt-free segmentation through hierarchical decoding and mask-guided attention. These studies demonstrate the effectiveness of foundation models and motivate more efficient prompt-free adaptation strategies.

\subsection{Probabilistic Anatomical Representation}

Prototype learning has been widely adopted for anatomical representation learning in medical image segmentation. Most methods represent each anatomical target using a deterministic prototype or feature center and perform similarity-based prediction. Although effective, such representations cannot adequately characterize the large intra-target variations and structural ambiguity commonly observed in medical images.

Recent advances in probabilistic representation learning provide a promising alternative. Representative methods, including Probabilistic U-Net~\cite{kohl2018probabilistic} and PHISeg~\cite{baumgartner2019phiseg}, model latent probability distributions to capture anatomical uncertainty. Subsequent studies further exploit Gaussian embeddings, probabilistic prototype learning, and distributional representations for feature learning, uncertainty estimation, and probabilistic classification~\cite{yuan2025probabilistic}, demonstrating richer semantic descriptions than deterministic feature centers.

However, their potential for learning anatomical representations for automatic prompt generation in prompt-free SAM remains largely unexplored.

\section{Methodology}
\label{sec:method}

\subsection{Overview}

\begin{figure*}[t]
    \centering
    \includegraphics[width=\textwidth]{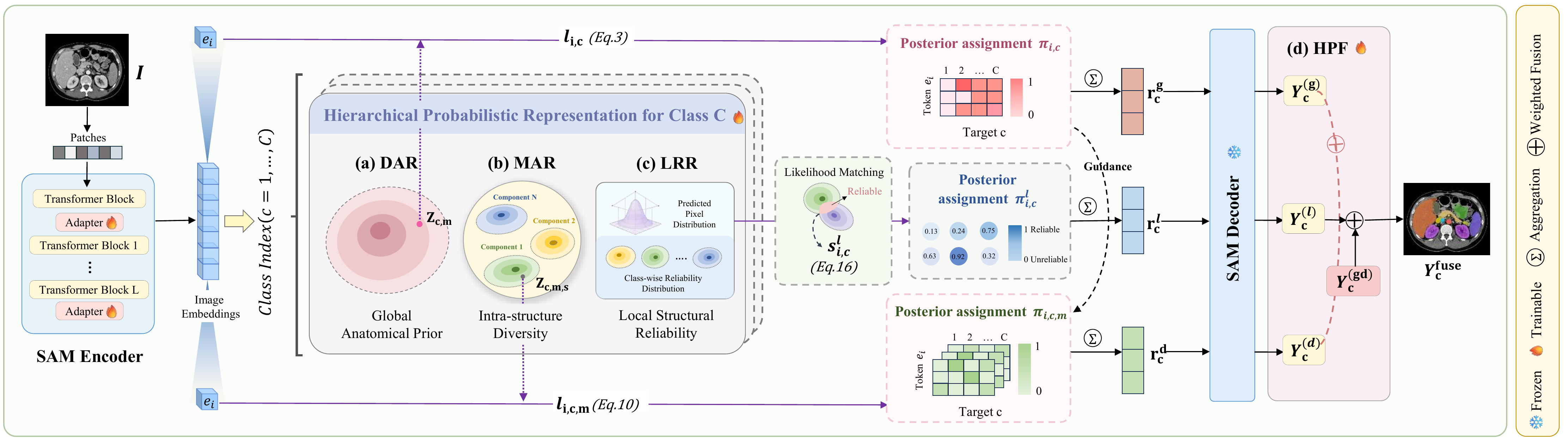} 
    \caption{
Overview of the proposed HPR-SAM framework.
Given image tokens extracted by the SAM encoder, HPR consists of four modules:
(a) Distributional Anatomical Representation (DAR) for global anatomical priors;
(b) Multi-component Anatomical Representation (MAR) for intra-structure diversity;
(c) Local Reliability Representation (LRR) for local structural reliability; and
(d) Hierarchical Prediction Fusion (HPF) for adaptive prediction fusion.
}
    \label{framework}
\end{figure*}

Fig.~\ref{framework} illustrates the overall framework of HPR. Unlike existing prompt-free SAM methods that mainly focus on prompt generator design, we argue that the quality of automatic prompts is fundamentally determined by the completeness of anatomical representations. Therefore, we formulate prompt-free medical image segmentation as a hierarchical probabilistic representation learning problem.

Given an input medical image $I$, the SAM image encoder~\cite{kirillov2023segment} $f_{\theta}(\cdot)$ extracts image embeddings $\mathbf{E}=f_{\theta}(I)\in\mathbb{R}^{N\times D}$, where $N=H'\times W'$ denotes the number of image tokens on the encoder feature grid and $D$ is the embedding dimension.

Based on $\mathbf{E}$, HPR learns three complementary probabilistic representations, denoted as $\mathbf{R}=\{\mathbf{R}^{g}, \mathbf{R}^{d},\mathbf{R}^{l}\}$. Specifically, $\mathbf{R}^{g}$ models global anatomical priors, $\mathbf{R}^{d}$ captures intra-structure anatomical diversity, and $\mathbf{R}^{l}$ estimates local structural reliability. These three representations are detailed in Sections~\ref{sec:global}, \ref{sec:diversity}, and \ref{sec:local}, respectively.

The learned representations are subsequently projected into SAM-compatible dense and sparse prompts and decoded by the original SAM mask decoder~\cite{kirillov2023segment}, producing three prediction maps $Y^{(g)}$, $Y^{(d)}$, and $Y^{(l)}$, corresponding to the three representation branches. Finally, a Hierarchical Prediction Fusion (HPF) module, detailed in Section~\ref{sec:fusion}, adaptively fuses these prediction maps to produce the final segmentation $Y^{fuse}$.

\subsection{Global Anatomical Prior Modeling}
\label{sec:global}

As the first component of a complete anatomical representation, global anatomical priors capture stable semantic characteristics shared across different patients and imaging conditions. Existing prompt-free methods often represent each anatomical target using a deterministic prototype or semantic token, which compresses complex anatomical information into a single point representation. Such deterministic representations are insufficient to describe the uncertainty and variability of anatomical structures. Motivated by probabilistic representation learning~\cite{kohl2018probabilistic,baumgartner2019phiseg}, we introduce the Distributional Anatomical Representation (DAR) module to address this limitation, which models the global prior of each anatomical target as a learnable probabilistic distribution and infers a global anatomical representation through posterior-guided token aggregation.

Specifically, DAR models each anatomical target $c\in\{1,\cdots,C\}$ as a Gaussian distribution:
\begin{equation}
\mathcal{G}_c=(\boldsymbol{\mu}_c,\boldsymbol{\Sigma}_c),
\qquad
q_c(z)=\mathcal{N}(\boldsymbol{\mu}_c,\boldsymbol{\Sigma}_c),
\end{equation}
where $\mathcal{N}(\cdot,\cdot)$ denotes a Gaussian distribution, $\boldsymbol{\mu}_c$ denotes the semantic center of anatomical target $c$, and $\boldsymbol{\Sigma}_c$ models the uncertainty of the corresponding anatomical prior.

Rather than directly using the Gaussian distribution as the global representation, DAR samples multiple latent anatomical hypotheses from the learned distribution:
\begin{equation}
z_{c,m}
=
\boldsymbol{\mu}_c
+
\boldsymbol{\Sigma}_c^{1/2}\boldsymbol{\epsilon}_m,
\qquad
\boldsymbol{\epsilon}_m
\sim
\mathcal{N}(\mathbf{0},\mathbf{I}_D),
\end{equation}
where $m$ indexes the sampled latent hypothesis, $\boldsymbol{\epsilon}_m$ is a standard Gaussian noise vector, $\mathbf{I}_D$ denotes the $D\times D$ identity matrix, and $\boldsymbol{\Sigma}_c^{1/2}$ denotes the matrix square root of the covariance matrix.

For each image token $e_i$, DAR estimates its compatibility with anatomical target $c$ by aggregating similarities over all sampled hypotheses:
\begin{equation}
\ell_{i,c}
=
\tau
\log
\sum_m
\exp
\left(
\frac{
\operatorname{sim}(e_i,z_{c,m})
}{\tau}
\right),
\end{equation}
where $\operatorname{sim}(\cdot,\cdot)$ denotes cosine similarity and $\tau$ is the temperature parameter.

The token-to-target posterior assignment is then computed as:
\begin{equation}
\pi_{i,c}
=
\frac{\exp(\ell_{i,c})}
{\sum_{k=1}^{C}\exp(\ell_{i,k})}.
\end{equation}
The resulting posterior reflects the probability that image token $e_i$ belongs to anatomical target $c$ under the learned probabilistic prior.

Finally, DAR aggregates image features according to the posterior assignment, yielding the global anatomical representation:
\begin{equation}
\mathbf{r}^{g}_{c}
=
\sum_{i=1}^{N}
\pi_{i,c}e_i.
\end{equation}

\subsection{Intra-structure Diversity Modeling}
\label{sec:diversity}

As the second component of a complete anatomical representation, intra-structure anatomical diversity characterizes multiple stable semantic patterns within the same anatomical target. Although the global anatomical prior learned by DAR captures stable semantic characteristics and models patient-level uncertainty using a single probabilistic distribution, all latent hypotheses are sampled from the same distribution, making it difficult to explicitly represent diverse anatomical sub-patterns within an anatomical target. Motivated by hierarchical probabilistic modeling~\cite{kohl2019hierarchical}, we introduce the Multi-component Anatomical Representation (MAR) module, which models each anatomical target using multiple probabilistic components. Each component is expected to capture one stable anatomical sub-pattern, and the diversity-aware representation is learned through hierarchical posterior-guided feature aggregation.

Specifically, MAR represents each anatomical target $c\in\{1,\cdots,C\}$ as a set of $M$ probabilistic components:
\begin{equation}
\mathcal{D}_c=
\{
\mathcal{G}_{c,1},
\mathcal{G}_{c,2},
\cdots,
\mathcal{G}_{c,M}
\},
\end{equation}
where each probabilistic component
\begin{equation}
\mathcal{G}_{c,m}
=
(
\boldsymbol{\mu}_{c,m},
\boldsymbol{\Sigma}_{c,m}
)
\end{equation}
models a distinct anatomical sub-pattern. The corresponding mixture distribution is formulated as
\begin{equation}
q_c(z)
=
\sum_{m=1}^{M}
\omega_{c,m}
\mathcal{N}
(
\boldsymbol{\mu}_{c,m},
\boldsymbol{\Sigma}_{c,m}
),
\end{equation}
where $\omega_{c,m}$ denotes the learnable mixture weight of the $m$-th probabilistic component.

Similar to DAR, MAR samples multiple latent anatomical hypotheses from each probabilistic component:
\begin{equation}
z_{c,m,s}
=
\boldsymbol{\mu}_{c,m}
+
\boldsymbol{\Sigma}_{c,m}^{1/2}
\boldsymbol{\epsilon}_{s},
\qquad
\boldsymbol{\epsilon}_{s}
\sim
\mathcal{N}
(
\mathbf0,
\mathbf I
),
\end{equation}
where $s$ indexes the sampled latent hypothesis.

For each image token $e_i$, MAR first estimates its compatibility with the $m$-th probabilistic component by aggregating similarities over all sampled hypotheses:
\begin{equation}
\ell_{i,c,m}
=
\tau
\log
\sum_{s}
\exp
\left(
\frac{
\operatorname{sim}
(
e_i,
z_{c,m,s}
)
}
{\tau}
\right).
\end{equation}

Based on the global posterior assignment $\pi_{i,c}$ inferred by DAR, MAR further computes the component-level posterior assignment as
\begin{equation}
\pi_{i,c,m}
=
\pi_{i,c}
\cdot
\operatorname{Softmax}_{m}
(
\ell_{i,c,m}
).
\end{equation}
The resulting component-level posterior reflects the probability that image token $e_i$ belongs to the $m$-th probabilistic component of anatomical target $c$.

Each probabilistic component independently aggregates image features:
\begin{equation}
\mathbf r_{c,m}
=
\sum_{i=1}^{N}
\pi_{i,c,m}
e_i.
\end{equation}

Finally, MAR integrates all component representations to obtain the diversity-aware anatomical representation:
\begin{equation}
\mathbf r_c^d
=
\sum_{m=1}^{M}
\omega_{c,m}
\mathbf r_{c,m}.
\end{equation}

\subsection{Local Structural Reliability Modeling}
\label{sec:local}

As the third component of a complete anatomical representation, local structural reliability distinguishes reliable local evidence from uncertain regions. Although DAR and MAR provide complementary global and diversity-aware anatomical semantics, they treat all image tokens equally during representation aggregation and therefore cannot explicitly suppress unreliable observations caused by ambiguous boundaries, low tissue contrast, or small anatomical structures. Inspired by probabilistic representation learning and mutual likelihood-based distribution matching~\cite{kohl2018probabilistic,baumgartner2019phiseg,yuan2025probabilistic}, we introduce the Local Reliability Representation (LRR) module to estimate token reliability and incorporate it into representation learning.

Given an image token $e_i$, LRR first parameterizes its local representation as a Gaussian distribution by predicting its mean and covariance:
\begin{equation}
\boldsymbol{\mu}_i=h_{\mu}(e_i),
\qquad
\boldsymbol{\Sigma}_i=h_{\Sigma}(e_i),
\end{equation}
where $h_{\mu}(\cdot)$ and $h_{\Sigma}(\cdot)$ are two lightweight prediction heads. The predicted mean $\boldsymbol{\mu}_i$ represents the local semantic embedding, while the covariance $\boldsymbol{\Sigma}_i$ characterizes its uncertainty.

To evaluate local reliability, LRR maintains a class-wise reliability distribution
\begin{equation}
\mathcal{P}_c=(\bar{\boldsymbol{\mu}}_c,\bar{\boldsymbol{\Sigma}}_c),
\end{equation}
where $\bar{\boldsymbol{\mu}}_c$ and $\bar{\boldsymbol{\Sigma}}_c$ denote the mean and covariance of anatomical target $c$. During training, these distributions are progressively updated using low-uncertainty observations through exponential moving average (EMA), serving as stable reliability references.

Instead of directly measuring feature similarity, LRR evaluates the agreement between the predicted token distribution and the corresponding reliability distribution using mutual likelihood:
\begin{equation}
s^{l}_{i,c}
=
-\frac{1}{2}
\left[
\frac{
\|\boldsymbol{\mu}_i-\bar{\boldsymbol{\mu}}_c\|^2
}
{
\boldsymbol{\Sigma}_i+\bar{\boldsymbol{\Sigma}}_c
}
+
\log(\boldsymbol{\Sigma}_i+\bar{\boldsymbol{\Sigma}}_c)
\right].
\end{equation}

The reliability-aware posterior is computed as
\begin{equation}
\pi^{l}_{i,c}
=
\frac{\exp(s^{l}_{i,c})}
{\sum_{k=1}^{C}\exp(s^{l}_{i,k})},
\end{equation}
where
\(
0<\pi^{l}_{i,c}<1
\)
and
\(
\sum_{c=1}^{C}\pi^{l}_{i,c}=1
\).
The resulting posterior therefore forms a normalized reliability-aware probability distribution over all anatomical targets, reflecting the confidence that image token $e_i$ provides reliable evidence for anatomical target $c$. It naturally serves as a reliability-aware weighting factor for subsequent feature aggregation.

Finally, LRR performs posterior-guided feature aggregation to construct the local reliability representation:
\begin{equation}
\mathbf{r}^{l}_{c}
=
\sum_{i=1}^{N}
\pi^{l}_{i,c}e_i.
\end{equation}

Unlike previous probabilistic matching methods that mainly employ posterior estimation for probabilistic classification, LRR uses the reliability-aware posterior to guide representation aggregation, reducing the influence of unreliable observations while emphasizing structurally consistent local evidence. The resulting representation complements DAR and MAR by providing reliability-aware local structural cues for the subsequent Hierarchical Prediction Fusion.

\subsection{Hierarchical Prediction Fusion}
\label{sec:fusion}

The three representation branches produce complementary prediction maps from different anatomical perspectives. Specifically, the global branch captures stable semantic priors, the diversity-aware branch preserves intra-structure variations, and the reliability-aware branch emphasizes reliable local evidence. To exploit their complementary strengths, we introduce a Hierarchical Prediction Fusion (HPF) strategy that adaptively integrates the prediction maps generated by the three branches.

Specifically, the learned global, diversity-aware, and reliability-aware representations are first projected into SAM-compatible dense and sparse prompts and decoded by the shared SAM mask decoder, producing three prediction maps $Y^{(g)}$, $Y^{(d)}$, and $Y^{(l)}$, respectively. Since different anatomical targets may benefit from different representation branches, HPF performs target-wise adaptive prediction fusion. For anatomical target $c$, the global and diversity-aware prediction maps are first combined:
\begin{equation}
Y^{(gd)}_{c}
=
(1-\alpha_c)Y^{(g)}_{c}
+
\alpha_c Y^{(d)}_{c},
\end{equation}
where $Y^{(gd)}_{c}$ denotes the intermediate prediction map fused from the global and diversity-aware branches, and $\alpha_c$ is a learnable target-specific fusion coefficient that balances the contribution of the diversity-aware prediction.

The reliability-aware prediction is then incorporated as local structural calibration:
\begin{equation}
Y^{fuse}_{c}
=
(1-\beta_c)Y^{(gd)}_{c}
+
\beta_cY^{(l)}_{c},
\end{equation}
where $\beta_c$ is a learnable target-specific reliability coefficient controlling the contribution of the reliability-aware prediction, and $Y^{fuse}_{c}$ denotes the final fused prediction for anatomical target $c$.

By hierarchically integrating complementary prediction maps from the three probabilistic representation branches, HPF enables each anatomical target to exploit stable global semantics, structural diversity, and reliable local evidence, yielding the final fused prediction $Y^{fuse}$.

\subsection{Training Objective}
\label{sec:objective}

Each prediction branch is supervised using the same segmentation loss:
\begin{equation}
\mathcal{L}_{seg}(Y,Y')
=
\mathcal{L}_{CE}(Y,Y')
+
\lambda_{Dice}\mathcal{L}_{Dice}(Y,Y'),
\end{equation}
where $Y$ denotes the ground-truth segmentation, and $Y'$ denotes either a branch prediction or the final fused prediction.

The branch supervision is computed as the average segmentation loss over the three representation branches:
\begin{equation}
\mathcal{L}_{branch}
=
\frac{1}{3}
\sum_{k\in\{g,d,l\}}
\mathcal{L}_{seg}(Y,Y^{(k)}),
\end{equation}
where $k\in\{g,d,l\}$ indexes the DAR, MAR, and LRR branches, respectively, and $Y^{(k)}$ denotes the corresponding prediction map.

The final fused prediction is supervised using the same segmentation loss:
\begin{equation}
\mathcal{L}_{fuse}
=
\mathcal{L}_{seg}(Y,Y^{fuse}),
\end{equation}
where $Y^{fuse}$ denotes the final prediction produced by Hierarchical Prediction Fusion (HPF).

Finally, the overall training objective is formulated as
\begin{equation}
\mathcal{L}_{total}
=
\mathcal{L}_{fuse}
+
\lambda_{branch}\mathcal{L}_{branch}.
\end{equation}

\section{Experiment} \label{sec:exp}

\subsection{Experimental Setup}

\noindent\textbf{Datasets.}
We conduct experiments on the Synapse medical image segmentation benchmark~\cite{landman2015miccai} and further evaluate the generalization capability of the proposed framework on LA~\cite{xiong2021global} and PROMISE12~\cite{litjens2014evaluation} under few-shot settings.

The Synapse dataset, derived from the MICCAI 2015 Multi-Atlas Abdomen Labeling Challenge~\cite{landman2015miccai}, contains 30 contrast-enhanced abdominal CT scans with voxel-wise annotations for the aorta, gallbladder, spleen, left kidney, right kidney, liver, pancreas, and stomach. Following previous SAM-based methods~\cite{zhang2023customized,cheng2024unleashing,zhong2025pg}, we adopt the standard split of 18 training and 12 testing cases.

For few-shot evaluation, LA contains 100 gadolinium-enhanced cardiac MRI volumes with left atrium annotations~\cite{xiong2021global}, where four labeled volumes are used for training without unlabeled data. PROMISE12 is a prostate MRI benchmark containing 50 annotated training cases~\cite{litjens2014evaluation}, where three labeled volumes are used under the same protocol.

\noindent\textbf{Implementation Details.}
For fair comparison, we follow the same data splits and evaluation protocols as previous SAM-based methods. All input slices are resized to $224\times224$ with standard data augmentation. The framework is implemented in PyTorch and trained on a single NVIDIA A40 GPU. We adopt ViT-B SAM with LoRA rank 4 following SAMed, and optimize the model using AdamW for 200 epochs. 

\begin{table*}[!t]
\centering
\scriptsize
\caption{Comparison with state-of-the-art methods on the Synapse CT dataset under the fully supervised setting. Bold values indicate the best performance among SAM-based methods. Mean Dice denotes the average Dice coefficient, and HD95 denotes the 95th percentile Hausdorff Distance.}
\label{tab:comparison}
\renewcommand{\arraystretch}{1.18}
\setlength{\tabcolsep}{4.2pt}
\resizebox{\textwidth}{!}{
\begin{tabular}{lcccccccccc}
\toprule
\textbf{Method} 
& \textbf{Spleen} & \textbf{Kidney(R)} & \textbf{Kidney(L)}
& \textbf{Gallbladder} & \textbf{Liver} & \textbf{Stomach}
& \textbf{Aorta} & \textbf{Pancreas}
& \textbf{Mean Dice[\%]}$\uparrow$ & \textbf{HD95[mm]}$\downarrow$ \\
\midrule
TransUnet~\cite{chen2021transunet} & 87.23 & 63.13 & 81.87 & 77.02 & 94.08 & 55.86 & 85.08 & 75.62 & 77.48 & 31.69 \\
SwinUnet~\cite{cao2022swin} & 85.47 & 66.53 & 83.28 & 79.61 & 94.29 & 56.58 & 90.66 & 76.60 & 79.13 & 21.55 \\
TransDeepLab~\cite{azad2022transdeeplab} & 86.04 & 69.16 & 84.08 & 79.88 & 93.53 & 61.19 & 89.00 & 78.40 & 80.16 & 21.25 \\
DAE-Former~\cite{azad2023dae} & 88.96 & 72.30 & 86.08 & 80.88 & 94.98 & 65.12 & 91.94 & 79.19 & 82.43 & 17.46 \\
MERIT~\cite{rahman2024multi} & 92.01 & 84.85 & 87.79 & 74.40 & 95.26 & 85.38 & 87.71 & 71.81 & 84.90 & 13.22 \\
\midrule
AutoSAM~\cite{hu2023efficiently} & 80.54 & 80.02 & 79.66 & 41.37 & 89.24 & 61.14 & 82.56 & 44.22 & 62.08 & 27.56 \\
SAM Adapter~\cite{chen2023sam} & 83.68 & 79.00 & 79.02 & 57.49 & 92.68 & 69.48 & 77.93 & 43.07 & 72.80 & 33.08 \\
SAMed~\cite{zhang2023customized} & 87.33 & 80.10 & 82.75 & 70.24 & 93.37 & 73.62 & 86.99 & 67.64 & 80.26 & 28.89 \\
H-SAM~\cite{cheng2024unleashing} & 92.34 & 85.99 & 87.71 & 69.65 & 95.20 & 86.27 & 87.53 & 72.53 & 84.65 & \textbf{7.29} \\
PG-SAM~\cite{zhong2025pg} & 93.12 & 84.57 & 87.93 & 73.26 & 95.40 & 86.62 & 87.87 & 71.49 & 84.79 & 7.61 \\
\midrule
\rowcolor[gray]{0.95}
\textbf{HPR-SAM (Ours)} & 91.38 & 85.92 & 87.76 & \textbf{73.31} & \textbf{95.58} & 84.25 & \textbf{89.43} & \textbf{73.10} & \textbf{85.09} & 11.94 \\
\bottomrule
\end{tabular}}
\end{table*}

\begin{table}[t]
\centering
\small
\caption{Comparison with representative segmentation methods under few-shot settings on the LA and PROMISE12 datasets. Mean Dice denotes the average Dice coefficient.}
\label{tab:merged_prompt_results}
\renewcommand{\arraystretch}{1.15}
\setlength{\tabcolsep}{12pt}
\begin{tabular}{lc}
\toprule
\textbf{Method} & \textbf{Mean Dice [\%]}$\uparrow$ \\
\midrule
\multicolumn{2}{l}{\textit{LA dataset~\cite{xiong2021global} (4 labeled scans, 0 unlabeled)}} \\
\midrule
nnUNet~\cite{isensee2021nnu} & 64.02 \\
AutoSAM~\cite{hu2023efficiently} & 74.73 \\
SAM Adapter~\cite{chen2023sam} & 82.79 \\
SAMed~\cite{zhang2023customized} & 82.27 \\
H-SAM~\cite{cheng2024unleashing} & 83.94 \\
\rowcolor[gray]{0.95}
\textbf{HPR-SAM (Ours)} & \textbf{84.65} \\
\midrule
\multicolumn{2}{l}{\textit{PROMISE12 dataset~\cite{litjens2014evaluation} (3 labeled scans, 0 unlabeled)}} \\
\midrule
AutoSAM~\cite{hu2023efficiently} & 68.40 \\
SAM Adapter~\cite{chen2023sam} & 75.45 \\
SAMed~\cite{zhang2023customized} & 78.55 \\
H-SAM~\cite{cheng2024unleashing} & 76.95 \\
\rowcolor[gray]{0.95}
\textbf{HPR-SAM (Ours)} & \textbf{81.26} \\
\bottomrule
\end{tabular}
\end{table}

\subsection{Comparisons with State-of-the-art Methods}

As shown in Table I, HPR-SAM achieves the highest Mean Dice on the Synapse CT dataset. Compared with conventional fully supervised methods, HPR-SAM achieves a Mean Dice of 85.09\%, outperforming TransUnet (+7.61\%), SwinUnet (+5.96\%), TransDeepLab (+4.93\%), DAE-Former (+2.66\%), and MERIT (+0.19\%). These results demonstrate the effectiveness of HPR-SAM while preserving the prompt-free SAM paradigm.

Compared with existing SAM-based methods, HPR-SAM consistently improves segmentation accuracy, surpassing AutoSAM (+23.01\%), SAM Adapter (+12.29\%), SAMed (+4.83\%), H-SAM (+0.44\%), and PG-SAM (+0.30\%) in Mean Dice. It also achieves the best Dice scores on the liver (95.58\%), aorta (89.43\%), and pancreas (73.10\%). Compared with PG-SAM, HPR-SAM further improves the Dice of the right kidney (+1.35\%), gallbladder (+0.05\%), liver (+0.18\%), aorta (+1.56\%), and pancreas (+1.61\%), indicating that hierarchical probabilistic representation learning generates more effective prompts and captures richer anatomical semantics and local structural details.

Although HPR-SAM does not achieve the lowest HD95 among SAM-based methods, its HD95 of 11.94 mm substantially outperforms AutoSAM (-15.62 mm), SAM Adapter (-21.14 mm), and SAMed (-16.95 mm), while remaining lower than several conventional fully supervised methods. H-SAM and PG-SAM achieve lower HD95 values, likely benefiting from additional decoder-side refinement for boundary enhancement. In contrast, HPR-SAM improves automatic prompt generation through hierarchical probabilistic representations while still achieving the highest Mean Dice and competitive boundary localization, demonstrating the effectiveness of anatomical representation learning for prompt-free SAM.

Beyond Synapse, we further evaluate HPR-SAM on the LA and PROMISE12 datasets under few-shot settings, as summarized in Table~\ref{tab:merged_prompt_results}. Using only four labeled scans on LA and three labeled scans on PROMISE12 without unlabeled data, HPR-SAM consistently achieves the best performance. On LA, HPR-SAM achieves a Mean Dice of 84.65\%, outperforming nnUNet (+20.63\%), AutoSAM (+9.92\%), SAM Adapter (+1.86\%), SAMed (+2.38\%), and H-SAM (+0.71\%). On PROMISE12, it achieves a Mean Dice of 81.26\%, surpassing AutoSAM (+12.86\%), SAM Adapter (+5.81\%), SAMed (+2.71\%), and H-SAM (+4.31\%). These consistent improvements on both cardiac and prostate MRI datasets demonstrate the strong generalization capability of HPR-SAM. By jointly modeling global anatomical priors, intra-structure diversity, and local structural reliability, HPR-SAM learns more transferable anatomical representations for prompt generation across imaging modalities under limited supervision.

\subsection{Qualitative Results}

\begin{figure*}[!t]
    \centering
    \includegraphics[width=0.8\textwidth]{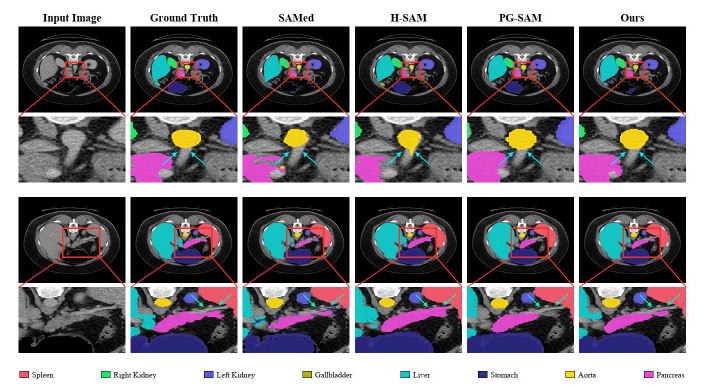}  
    \caption{
   Qualitative comparison of segmentation results produced by different methods on the Synapse dataset, with particular focus on the aorta and pancreas.
    }
    \label{quante}
\end{figure*}

As shown in Fig.~\ref{quante}, we compare HPR-SAM with representative prompt-free SAM-based medical segmentation methods, including SAMed, H-SAM, and PG-SAM, on the Synapse dataset, focusing on the aorta and pancreas. These two anatomical targets are selected due to their boundary ambiguity, anatomical variation, and close proximity to surrounding tissues in abdominal CT images, making them prone to local mis-segmentation and incomplete delineation.

For the aorta, HPR-SAM produces results more consistent with the ground truth, especially in the enlarged regions. As indicated by the cyan arrows, SAMed shows obvious under-segmentation, H-SAM introduces local errors around the aorta, and PG-SAM still exhibits contour mismatch despite relatively complete predictions. In contrast, HPR-SAM better preserves the location and shape of the aorta, suggesting that hierarchical anatomical representation provides more reliable constraints for automatic prompt generation.

For the pancreas, segmentation is particularly challenging due to its elongated shape, low contrast, and indistinct boundaries. As indicated by the light green arrows, SAMed misses both ends of the pancreas, H-SAM over-segments surrounding tissues, and PG-SAM accurately delineates the left part but fails to fully cover the right part. In contrast, HPR-SAM produces a more complete and anatomically consistent segmentation, better balancing structural completeness and boundary precision.

\subsection{Ablation Study}
To validate the effectiveness of the proposed hierarchical probabilistic representation learning framework, we conduct ablation studies on the Synapse dataset, as summarized in Table~\ref{tab:ablation_components}. The baseline model without DAR, MAR, and LRR achieves a Mean Dice of 74.86\% and an HD95 of 27.98~mm.

Introducing DAR improves the Mean Dice to 82.74\% (+7.88\%) and reduces HD95 to 20.52~mm (-7.46~mm), demonstrating that probabilistic global priors provide more stable anatomical guidance than deterministic representations. MAR further increases the Mean Dice to 82.88\% (+8.02\%) and reduces HD95 to 18.19~mm (-9.79~mm), indicating its effectiveness in modeling intra-structure variations. LRR alone improves the Mean Dice to 82.30\% (+7.44\%) but increases HD95 to 29.61~mm (+1.63~mm), suggesting that local reliability modeling is less effective without support from global or diversity-aware representations.

Combining components further improves performance. DAR+MAR achieves 83.36\% Mean Dice (+8.50\%) and 17.82~mm HD95 (-10.16~mm), demonstrating the complementarity between global priors and intra-structure diversity. Adding LRR to DAR or MAR further improves Mean Dice to 84.32\% (+9.46\%) and 84.21\% (+9.35\%), respectively, indicating that reliability modeling is more effective when supported by complementary representations.

Finally, integrating all three components achieves the best performance, with a Mean Dice of 85.09\% (+10.23\%) and an HD95 of 11.94~mm (-16.04~mm). These results show that DAR, MAR, and LRR provide complementary probabilistic representations, whose hierarchical integration improves both region accuracy and boundary localization.

\begin{table}[H]
\centering
\caption{Ablation study of different components.}
\label{tab:ablation_components}
\begin{tabular}{ccc|cc}
\hline
\textbf{DAR} & \textbf{MAR} & \textbf{LRR} & \textbf{Mean Dice (\%)} & \textbf{Mean HD95 (mm)} \\
\hline
$\times$ & $\times$ & $\times$ & 74.86 & 27.98 \\
$\checkmark$ & $\times$ & $\times$ & 82.74 & 20.52 \\
$\times$ & $\checkmark$ & $\times$ & 82.88 & 18.19 \\
$\times$ & $\times$ & $\checkmark$ & 82.30 & 29.61 \\
$\checkmark$ & $\checkmark$ & $\times$ & 83.36 & 17.82 \\
$\checkmark$ & $\times$ & $\checkmark$ & 84.32 & 18.08 \\
$\times$ & $\checkmark$ & $\checkmark$ & 84.21 & 17.74 \\
$\checkmark$ & $\checkmark$ & $\checkmark$ & \textbf{85.09} & \textbf{11.94} \\
\hline
\end{tabular}
\end{table}

\section{Conclusion}

In this paper, we propose HPR, a hierarchical probabilistic representation learning framework for prompt-free SAM-based medical image segmentation. Instead of directly optimizing prompt generation, HPR learns anatomically complete representations by jointly modeling global anatomical priors, intra-structure diversity, and local structural reliability. Specifically, DAR captures global anatomical priors, MAR captures intra-structure diversity, and LRR emphasizes reliable local evidence. These complementary representations are projected into SAM-compatible prompts and integrated through Hierarchical Prediction Fusion (HPF). Experiments on the Synapse, LA, and PROMISE12 datasets demonstrate the effectiveness of HPR, achieving state-of-the-art performance on Synapse and under few-shot settings on LA and PROMISE12. These results suggest that hierarchical probabilistic representation learning provides an effective representation paradigm for prompt-free SAM-based medical image segmentation.

\bibliographystyle{IEEEtran}
\bibliography{references}
\end{document}